\documentclass[sigconf]{acmart}
\usepackage{multirow}
\usepackage{listings}
\usepackage{xcolor}

\usepackage{epigraph} 
\AtBeginDocument{%
  }

\setcopyright{acmlicensed}
\copyrightyear{2018}
\acmYear{2018}
\acmDOI{XXXXXXX.XXXXXXX}
\acmConference[Conference acronym 'XX]{Make sure to enter the correct
  conference title from your rights confirmation email}{June 03--05,
  2018}{Woodstock, NY}
\acmISBN{978-1-4503-XXXX-X/2018/06}

\begin{document}

\title{Does Language Model Understand Language?}

%




\author{Suvojit Acharjee}
\authornote{All authors contributed equally to this research.}
\affiliation{%
  \institution{Institute of Engineering and Management}
  \city{Saltlake, Kolkata}
  \state{West Bengal}
  \country{India}
}
\email{acharjeesuvo@gmail.com}
\orcid{0000-0002-7540-4777}

\author{Utathya Aich}
\authornotemark[1]
\affiliation{%
  \institution{Jadavpur University}
  \city{Kolkata}
  \state{West Bengal}
  \country{India}
}
\email{us4decaich@gmail.com}

\author{Asfak Ali}
\authornotemark[1]
\affiliation{%
  \institution{Jadavpur University}
  \city{Kolkata}
  \state{West Bengal}
  \country{India}
}
\email{asfakali.etce@gmail.com}

\renewcommand{\shortauthors}{acharjee et al.}

\begin{abstract}
Despite impressive advances in natural language generation and understanding, modern language models (LMs) still struggle with fine-grained linguistic phenomena such as tense, negation, voice, and modality which are the elements central to effective human communication. In the context of the United Nations Sustainable Development Goal 4 (Quality Education), where linguistic clarity and contextual precision are critical, the deployment of LMs in educational technologies demands careful scrutiny. As LMs are increasingly powering applications like tutoring systems, automated grading, and translation, their alignment with human linguistic interpretation becomes essential for equitable and effective learning. In this study, we conduct a systematic evaluation of state-of-the-art language models across these challenging contexts in both English and Bengali. To ensure a structured and rigorous assessment, we introduce the \textit{Preferred Route for Evaluation of Cognitive Inference in Systematic Environments} (PRECISE) guidelines. Our proposed Linguistic Understanding and Cognitive Inference Dataset (LUCID), composed of carefully crafted sentence pairs with a 70:30 ratio between English and Bengali, specifically challenges these models on critical aspects of language comprehension, including negation, tense, voice, and modal variations. We assess the performance of cutting-edge models including MISTRAL-SABA-24B, LLaMA-4-Scout-17B, LLaMA-3.3-70B-Versatile, Gemma2-9B, and Compound-Beta using standard metrics like Pearson correlation, Spearman correlation, and Mean Absolute Error (MAE), as well as a novel, linguistically inspired metric—the Human Calibration Envelope (HCE) accuracy. The HCE accuracy measures how often model predictions fall within one standard deviation of the mean human rating, thus capturing human-like tolerance for variability in language interpretation. Our findings highlight {Compound-Beta} as the most balanced model, consistently achieving high correlations and low MAEs across diverse language conditions. It records the highest Pearson correlation in English (0.8307) and demonstrates robust performance on mixed-language data (Pearson: 0.7981; HCE: 0.6290), indicating a strong alignment with human judgments in both monolingual and cross-lingual scenarios.

\end{abstract}



\maketitle

\section{Introduction}

\epigraph{"Learning Language is Harder Than You Think"}{\textit{Gary Marcus  \\ 2022}}

Language is a powerful tool that shapes every aspect of our lives. From sharing our thoughts and emotions to forging connections with others, language serves as the cornerstone of communication and understanding. It plays a vital role in the preservation of our culture and the promotion of creativity, learning, and growth. Language allows us to influence and inspire, build relationships and communities, and navigate the complex tapestry of human interaction.  Quality education (SDG 4) hinges not only on the availability of learning materials but also on their linguistic clarity, contextual accuracy, and accessibility across diverse linguistic populations. The growing integration of language models (LMs) in educational technologies—from intelligent tutoring systems to automated grading and translation services—makes their linguistic fidelity and cognitive alignment with human interpretation increasingly consequential.

With the advancement of machine learning and artificial intelligence, we have developed large language models capable of analyzing and generating human-like text. These models learn statistical patterns from vast amounts of human-generated content, emulating the ways we use language to communicate and connect. However, there are differing viewpoints on how this learning process aligns with human language acquisition. The tabula rasa theory, advocated by John Locke \cite{tabularosa}, posits that humans are born as blank slates, gaining knowledge through experience and perception—much like these models. In contrast, the nativist theory, championed by thinkers such as Plato and Kant, suggests that we possess innate knowledge of the world. \cite{Marcus_2022} This idea was further developed by the linguist Noam Chomsky, who argued that humans have an inherent understanding of universal grammar, forming a foundational framework for language \cite{chomsky1991universal}. These debates highlight the ongoing discussion about how language, as a fundamental part of our lives, is acquired and understood.

The challenge of learning statistical patterns from examples without comprehending the underlying principles makes any system inherently brittle and highly sensitive to outliers and noise. While large neural networks, with their sheer size and capacity, can mitigate this brittleness to some extent by absorbing variations in the data, they do not provide a complete solution. These networks operate through statistical pattern matching in high-dimensional input spaces, which means they remain susceptible to the same fundamental limitations as simpler models. Without a deeper understanding of the underlying structure and causal relationships in the data, such models can struggle with generalization and robustness when faced with unfamiliar or noisy inputs. 

\begin{figure}
    \centering
    \includegraphics[width=1\linewidth]{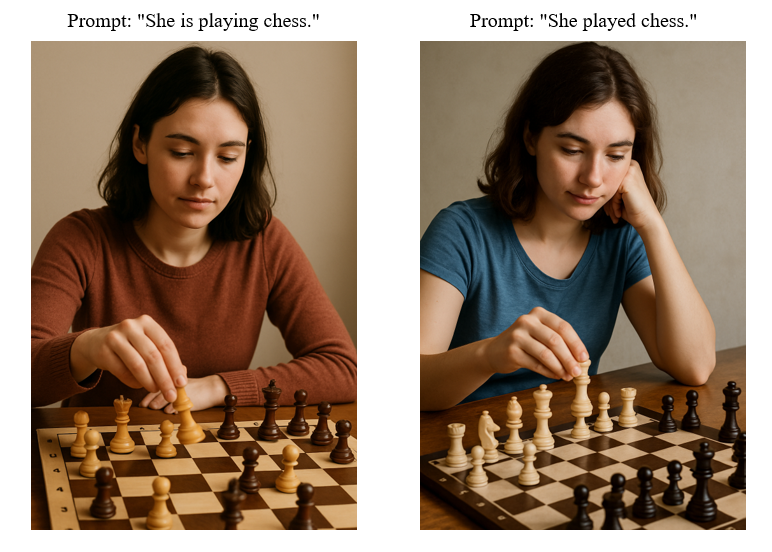}
    \caption{ChatGPT generated image for two different prompts.}
    \label{fig1}
\end{figure}

 Language models (LMs) are a prime example of such large deep neural networks and thus inherit these same limitations. They are prone to issues like hallucination, bias, challenges in generalization, and a lack of interpretability, all of which can undermine their reliability. However, despite these inherent challenges, recent developments suggest that LMs are nonetheless being rapidly and widely adopted across a variety of fields, driven by their impressive capabilities in capturing patterns and generating human-like text. Large language models (LLM) have quietly become integral to various applications, such as search engines, document summarization, code generation, and chat-based customer support systems. Meanwhile, vision-language models (VLM) have revolutionized tasks like image generation, captioning, and even multimodal reasoning, expanding the ways in which computers can understand and create visual content. These advancements provide users with intuitive natural language interfaces, making complex tasks more accessible and interactive. For instance, language models are now used to generate realistic images from text descriptions, power advanced translation systems, and even assist in scientific research and creative writing. Given this growing reliance on LMs, it becomes crucial to rigorously evaluate their performance across diverse scenarios, including different languages, cultures, and subject areas, to ensure that these models are robust, reliable, and aligned with human needs and expectations. For example, we asked ChatGPT to generate an image from two separate prompts in two different sessions:
\begin{lstlisting}
Prompt1: Generate an image: "She @h@is playing@ chess." 
Prompt2: Generate an image: "She @h@played@ chess."
\end{lstlisting}
The images generated by ChatGPT, as shown in \autoref{fig1}, failed to capture the temporal difference between the prompts, as both images depict her actively playing chess in the present moment. However, the prompts clearly distinguish between a situation where she is currently playing chess and another where she has already played chess. This is highlighting how the current models still struggle to accurately represent subtle nuances like tense and temporal context in language descriptions. In this article, we delve into the ability of various LMs to accurately distinguish subtle differences in input sentences and handle a range of linguistic hedges. Understanding these capabilities is crucial, as their performance directly influences a wide range of downstream applications, including multimedia generation systems, computer vision, robot perception, navigation, path planning, and many other domains that rely on language models as their foundational interface or control mechanism. Our main \textbf{contributions} are as follows:
\begin{enumerate}
    \item A comprehensive evaluation of LMs based on their ability to classify and differentiate between pairs of sentences with either similar structure but different meanings or similar meanings but different structures.
    \item The creation of a multilingual dataset called Linguistic Understanding and Cognitive Inference Dataset (LUCID) \footnote{https://github.com/Luciddataset/LUCID}, containing pairs of sentences that test various linguistic nuances, designed to challenge and benchmark the models’ understanding of subtleties in language. The LUCID dataset comprises examples in two languages: English and Bengali.
    \item We have introduced Human Calibration Envelope (HCE) accuracy metric reflects a shift toward evaluating models not only by their raw performance but by their alignment with human cognitive tolerance and variability. This is particularly relevant to SDG 3 (Good Health and Well-being), as LLMs are increasingly deployed in contexts that affect mental health, such as conversational agents and educational support tools. By ensuring that models can correctly interpret sensitive linguistic cues like negation and modality, we contribute to safer, more supportive digital environments. The promotion of transparent, linguistically competent AI also supports the objectives of SDG 16 (Peace, Justice, and Strong Institutions), by improving fairness and accountability in automated decision-making systems used in legal, educational, and administrative contexts.

    \item A detailed analysis of how LMs handle tense, temporal context, and syntactic variations, shedding light on their limitations and strengths in nuanced language understanding.
    \item Recommendations and guidelines for improving LM performance in applications that rely on precise understanding of meaning and context, including search engines, conversational AI, multimedia generation systems, computer vision, and robot perception/navigation/path planning.
\end{enumerate}
By examining these aspects, we aim to provide a clearer picture of the current capabilities and limitations of language models in handling complex linguistic features. In the next section, we will review several related works in this area. The methodology and design of experiments are detailed in Section 3, followed by an analysis of the results presented in Section 4. Finally, Section 5 concludes the article with key findings and future directions.
\section{Literature Review}
The ability of LMs to synthesize well-structured phrases is impressive, but they often struggle to understand subtle distinctions in meaning. LMs are therefore sometime termed as "Stochastic Parrots" by Bender et al \cite{stochasticparrot}. In their Two Word Test (TWT), Nicholas Riccardi and Rutvik H. Desai demonstrated that humans excel at combining multiple words to form a single coherent concept, while LMs frequently fall short in this regard \cite{TWT}.

The TWT experiment involved creating a dataset of 1,768 pairs of noun-noun combinations. Some of these pairs represented meaningful concepts, such as "baby boy," while others were nonsensical combinations like "goat sky." Human volunteers rated these combinations on a scale of 0 –4 and also provided binary judgments of "sensible" or "nonsense" for each phrase.

LLMs such as GPT-3.5, Bard, and GPT-4 were also tasked with evaluating these same combinations. When their ratings were compared to those of human participants, the LLMs performed poorly overall. Notably, GPT-3.5 and Bard failed even in the binary judgments, misclassifying sensible and nonsensical pairs. However, GPT-4 showed a significant improvement in binary judgments compared to the earlier models, although it still lagged considerably behind human performance.

 In another experiment, Jang et al. \cite{negated} conducted a case study that clearly demonstrates how LM performance declines when handling negated prompts. While LMs typically perform well with direct, straightforward prompts—sometimes even surpassing human performance—their accuracy drops significantly when those prompts are negated, underscoring the models’ struggle to handle subtle logical transformations in language. Similarly, Arkoudas, through systematic evaluations, concludes that GPT-4 lacks genuine reasoning capabilities.

 Overall, these studies underscore the significant gap between language models' apparent fluency in generating language and their actual comprehension of meaning and context. While recent advancements have led to notable improvements, limitations remain—particularly in reasoning and handling negations. Moreover, although there are numerous studies evaluating the understanding of LMs in English, there is a noticeable scarcity of evaluations focusing on Indic languages. This article aims to address this gap by providing a systematic evaluation of LMs’ understanding not only in English but also in Bengali, thereby offering a more comprehensive perspective on their capabilities and limitations across languages.

 \section{Methodology}
 To evaluate the understanding of LMs, we employ a systematic, step-by-step approach that we refer to as the Preferred Route for Evaluation of Cognitive Inference in Systematic Environments (PRECISE) guidelines. This methodology not only provides a structured framework for assessing language models but can also be adapted to evaluate other intelligent systems. The steps of the PRECISE guidelines are outlined below:
 \begin{itemize}
     \item[\textbf{1.}] \textbf{Application Identification:}Identify the application that you wish to evaluate. In this article, the focus is on LMs.
     \item[\textbf{2.}] \textbf{Task Identification:} Identify the specific task for evaluation within your chosen application. In this article, we focus on evaluating the understanding of LMs.
    \item[\textbf{3.}]\textbf{Challenges Identification} Having identified the application and the specific task for evaluation, the next step is to pinpoint the challenging conditions that may arise during this task. For language models, this involves considering linguistic nuances that can be difficult to interpret correctly. We have identified four such conditions in the language:
        \begin{figure}
        \centering
        \includegraphics[width=1\linewidth]{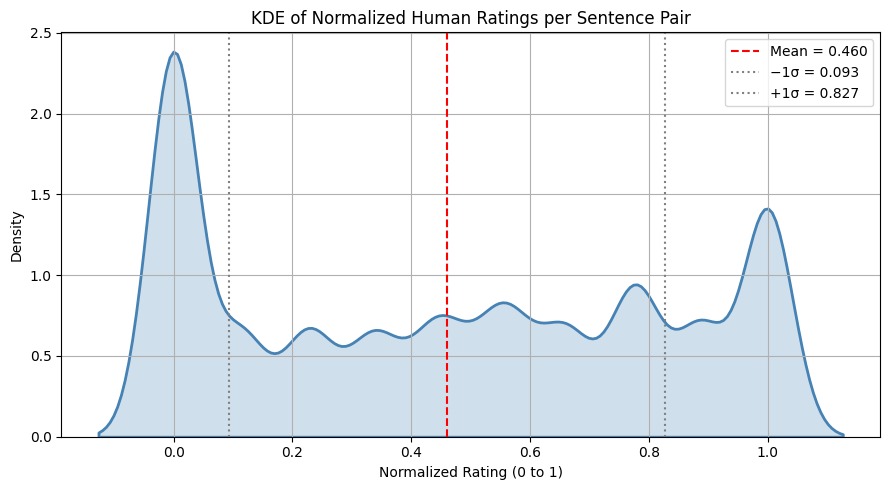}
        \caption{Distribution of average normalized similarity ratings (0–1 scale) for Bengali and English sentence pairs. Ratings were collected from human annotators on a 1–10 scale and then normalized.  The bimodal distribution suggests annotators often agreed on sentence pairs being either clearly similar or dissimilar.
}
        \label{figkde}
    \end{figure}
    \begin{itemize}
         \item \textbf{Change of speech:} When a sentence transitions from direct to indirect speech, it requires the model to understand not just the literal meaning of words, but also how context, pronouns, and verb tense shift in retelling a statement.
         \item \textbf{Tense:} The concept of tense introduces subtle but critical temporal differences between sentences that share the same structure, involving the same subject, object, and verbs.
         \item \textbf{Negation:} Another major challenge is that negation fundamentally changes the meaning of a sentence. Our literature review has already shown that LLMs often struggle with negated prompts. Therefore, it is crucial to evaluate how well LMs understand and handle negated sentences.
         \item \textbf{Modal verbs and ambiguity:} Modal verbs introduce ambiguity and subtle shifts in meaning, and their effect on LMs’ understanding must be thoroughly evaluated.
    \end{itemize}
    \item[\textbf{4.}] \textbf{Dataset Generation/Selection:} After identifying the challenging conditions, the next step is to select or create a dataset that will test the application for the specific task and, importantly, emphasize these challenging areas. Ideally, the dataset should consist of a diverse set of examples drawn primarily from these challenging linguistic constructs. For example, in this article, we have curated a dataset that specifically targets the ability of language models to handle changes in voice (direct vs indirect speech), tense variations (temporal differences), negation (affirmative vs negative sentences), and the usage of modal verbs (such as "can," "must," "should," etc.). This focused dataset ensures that our evaluation is robust and meaningful, highlighting the nuanced areas where language models often struggle.
    \begin{itemize}
        \item \textbf{Dataset Description:} The proposed Linguistic Understanding and Cognitive Inference Dataset (LUCID) comprises of carefully crafted sentence pairs with a 70:30 ratio between English and Bengali language. Each language set is further subdivided into several challenging linguistic constructs, including: negation-affirmative pairs, same-tense different-voice pairs, different-tense same-voice pairs, different - tense different-voice pairs, and ambiguous sentence pairs. This detailed categorization ensures comprehensive coverage of the nuanced linguistic features that test the cognitive inference abilities of the language models under evaluation pair. The structure of the LUCID dataset is detailed in \autoref{fig2}.

        \autoref{figkde} illustrates the distribution of normalized semantic similarity scores assigned by human annotators to sentence pairs in our dataset. Annotators rated each pair on a scale from 1 to 10 based on perceived semantic similarity; these scores were then linearly normalized to the range [0, 1]. The dataset comprises 30\% Bengali-Bengali and 70\% English-English sentence pairs, each evaluated by multiple annotators. The kernel density estimate (KDE) reflects the average normalized score per sentence pair, aggregated across annotators.
        
        The resulting distribution exhibits a distinctly bimodal shape, with prominent peaks near 0 and 1. This suggests that annotators consistently judged most sentence pairs as either clearly dissimilar or clearly similar, with relatively few ratings occupying the ambiguous mid-range. The mean normalized similarity is approximately 0.460, with a lower concentration of scores near 0.093 and a higher density near 0.827. This high inter-annotator agreement on extreme similarity values offers a reliable foundation for training and evaluating computational models of semantic similarity. 
    \end{itemize}
    \begin{figure*}
        \centering
        \includegraphics[width=1\linewidth]{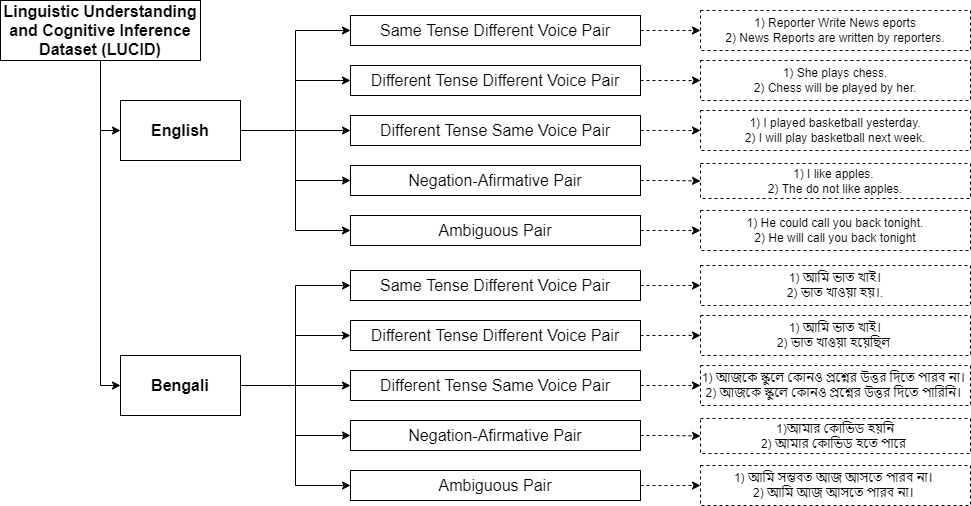}
        \caption{Structure of LUCID dataset}
        \label{fig2}
    \end{figure*}
    \item[\textbf{5.}] \textbf{Identification of State-of-the-Art Models/Algorithms:} The next step involves identifying and selecting state-of-the-art (SOTA) language models to evaluate their ability to understand subtle linguistic features relevant to this task. In this article, we have chosen a diverse and representative set of SOTA LMs for this evaluation, including MISTRAL-SABA-24B,  LLaMA-4-Scout-17B-16E,  LLaMA-3.3-70B-Versatile,  Gemma2-9B,  and Compound-Beta.
    \item[\textbf{6.}] \textbf{Inference Execution:} The subsequent step involves running inference on the chosen or generated dataset (LUCID) using the identified SOTA LMs. The outcomes from these inference runs are then carefully recorded for further analysis.
    \item[\textbf{7.}] \textbf{Analysis:} It is essential to conduct a thorough analysis of the outputs produced by the SOTA LMs. This analysis can be based on well-established evaluation metrics or on newly proposed metrics tailored to the specific nuances of the dataset and task. If a novel metric is introduced, it should be rigorously defined and clearly discussed to ensure transparency and replicability.

 \end{itemize}
    
\subsection{Human Calibration Envelope (HCE): A Robust Metric for Linguistic Alignment}

A central challenge in evaluating whether a language model ``understands'' language lies in quantifying the \textit{alignment} between human semantic judgments and model-produced similarity scores. Traditional metrics such as Pearson's $r$, Spearman's $\rho$, and absolute deviation (e.g., MAE) offer aggregate insights, but fail to account for the inherent \textit{variability} and \textit{subjectivity} in human judgments. Particularly across linguistic phenomena like negation or modality, human semantic perception varies across annotators.

To address this, we propose a novel metric: the \textbf{Human Calibration Envelope (HCE)}. Unlike scalar alignment measures, HCE evaluates whether a model's response falls within the empirically derived range of human variation for a given sentence pair.

\paragraph{Definition and Computation.} Let each sentence pair $i$ receive similarity ratings from $M$ human annotators, yielding a mean score $\bar{s}^{\text{human}}_i$ and standard deviation $\sigma^{\text{human}}_i$. We define the \textit{Human Calibration Envelope} for pair $i$ as:

\begin{equation}
    \mathcal{E}_i = \left[ \bar{s}^{\text{human}}_i - \sigma^{\text{human}}_i,\; \bar{s}^{\text{human}}_i + \sigma^{\text{human}}_i \right]
\end{equation}

A model's similarity score $s^{\text{LLM}}_i$ is considered \textit{calibrated} if it falls within this interval:

\begin{equation}
    \text{HCE}_i = 
    \begin{cases}
        1 & \text{if } s^{\text{LLM}}_i \in \mathcal{E}_i \\
        0 & \text{otherwise}
    \end{cases}
\end{equation}

The overall \textbf{HCE Accuracy} is computed as:

\begin{equation}
    \text{HCE}_{\text{accuracy}} = \frac{1}{N} \sum_{i=1}^N \text{HCE}_i
\end{equation}
 The HCE metric captures \textit{distribution-aware alignment}, recognizing that human ratings are not absolute.

\section{Result and Discussion}

The understanding of LMs is evaluated across three complementary dimensions. The first dimension assesses the degree of alignment between LM predictions and human ratings. The second dimension examines the quality of the reasoning explanations provided by the LMs. The third dimension analyzes the disagreement patterns between human ratings and model predictions to reveal potential biases and gaps in comprehension.
\begin{table*}[!t]
\centering
\caption{Performance Comparison of LLMs across English, Bengali, and Mixed Sentence Pairs.}
\label{tab:llm_language_split}

\begin{tabular}{lcccccccccc}
\toprule
\textbf{Model} & \textbf{Lang} & \textbf{Pearson} & \textbf{Spearman} & \textbf{MAE} ↓ & \textbf{HCE Acc.} ↑ \\
\midrule
\multirow{3}{*}{Mistral-SABA-24B \cite{jiang2024identifying}} 
  & English & 0.5404 & 0.5378 & 2.8883 & 0.4510 \\
  & Bengali & 0.4560 & 0.4312 & 3.3484 & 0.2727 \\
  & Mixed   & 0.5294 & 0.5227 & 2.9699 & 0.4194 \\
\midrule
\multirow{3}{*}{LLaMA-4-Scout-17B \cite{touvron2023llama}}
  & English & 0.4518 & 0.5118 & 3.5111 & 0.3529 \\
  & Bengali & 0.5709 & 0.5196 & 4.2574 & 0.0909 \\
  & Mixed   & 0.4689 & 0.4889 & 3.6435 & 0.3065 \\
\midrule
\multirow{3}{*}{LLaMA-3.3-70B-Versatile \cite{grattafiori2024llama}}
  & English & 0.8063 & 0.8703 & 2.7323 & 0.4510 \\
  & Bengali & 0.4215 & 0.7116 & 4.0035 & 0.0909 \\
  & Mixed   & 0.7583 & 0.8805 & 2.9579 & 0.3871 \\
\midrule
\multirow{3}{*}{Gemma2-9B \cite{team2024gemma}}
  & English & 0.8729 & 0.9284 & 1.7345 & 0.7843 \\
  & Bengali & 0.6857 & 0.7121 & 2.8017 & 0.4545 \\
  & Mixed   & 0.8272 & 0.8897 & 1.9239 & 0.7258 \\
\midrule
\multirow{3}{*}{Compound-Beta \cite{upadhye2019compound}}
  & English & 0.8307 & 0.8240 & 1.9838 & 0.6667 \\
  & Bengali & 0.6177 & 0.6533 & 2.9111 & 0.4545 \\
  & Mixed   & 0.7981 & 0.8257 & 2.1483 & 0.6290 \\
\bottomrule
\end{tabular}\label{tab:llm_similarity}
\end{table*}
\subsection{Quantification of Understanding}
Table~\ref{tab:llm_language_split} provides a detailed breakdown of model performance across three linguistic categories: English, Bengali, and Mixed (cross-lingual) sentence pairs. The evaluation is based on four complementary metrics: Pearson \cite{ahmed2018pearson} and Spearman \cite{doi:https://doi.org/10.1002/0470011815.b2a15150} correlations, which measure alignment with human similarity judgments; Mean Absolute Error (MAE), indicating prediction precision; and the Human Calibration Envelope (HCE) Accuracy, a linguistically motivated metric quantifying the frequency with which model predictions fall within one standard deviation of the mean human rating—thus capturing tolerance for intersubjective variation.
\begin{table*}[!t]
\centering
\caption{Average reasoning quality scores across prompt categories for different models, where \textbf{Same Tense Different Voice (STDV)}, \textbf{Different Tense Different Voice (DTDV)}, \textbf{Different Tense Same Voice (DTSV)}, \textbf{Negation- Affirmative (NA)}.}
\begin{tabular}{lccccccc}
\hline
\textbf{Model} & \textbf{Overall} & \textbf{STDV} & \textbf{DTDV} & \textbf{DTSV} & \textbf{NA} & \textbf{Ambiguous} & \textbf{Bengali} \\
\hline
Mistral-SABA-24B & 0.1808 & 0.1665 & 0.1114 & 0.1465 & 0.0711 & 0.3972 & 0.3170 \\
LLaMA-4-Scout-17B & 0.2200 & 0.2533 & 0.2213 & 0.1651 & 0.0401 & 0.3823 & 0.3317 \\
LLaMA-3.3-70B-Versatile & 0.1701 & 0.2065 & 0.1377 & 0.1280 & 0.0060 & 0.3693 & 0.2720 \\
Gemma2-9B & 0.1573 & 0.1664 & 0.0799 & 0.1301 & 0.0126 & 0.3510 & 0.3141 \\
Compound-Beta & 0.2000 & 0.2278 & 0.1868 & 0.1397 & 0.0527 & 0.3970 & 0.2913 \\
\hline
\end{tabular}

\label{tab:reasoning_quality}
\end{table*}
Across all language conditions, \textbf{Compound-Beta} emerges as the most balanced model, achieving consistently high correlations and low MAEs. In particular, it attains the highest English Pearson correlation (0.8307) and strong performance on mixed data (Pearson: 0.7981; HCE: 0.6290), indicating a strong alignment with human judgment across both monolingual and cross-lingual inputs. \textbf{Gemma2-9B} stands out for its exceptional performance in English (Pearson: 0.8729, MAE: 1.7345), and demonstrates remarkable calibration sensitivity, reaching the highest HCE accuracy across all languages (English: 0.7843; Mixed: 0.7258). This suggests that Gemma2-9B is particularly well-calibrated to human rating distributions, a crucial aspect when modeling interpretive nuances.

\textbf{LLaMA-3.3-70B-Versatile} performs competitively in English and mixed settings (Pearson: 0.8063 and 0.7583 respectively), but underperforms in Bengali (Pearson: 0.4215; HCE: 0.0909), highlighting challenges in adapting to low-resource or morphologically rich languages.
\textbf{LLaMA-4-Scout-17B}, while showing moderate performance in English and mixed settings, exhibits a notable drop in HCE accuracy for Bengali (0.0909) despite a relatively high Pearson correlation (0.5709), suggesting inconsistencies between ranking ability and tolerance of human disagreement.

Finally, \textbf{Mistral-SABA-24B} delivers stable, mid-tier performance across all languages but does not lead on any single metric, making it a competent generalist rather than a specialized performer. These findings reinforce two key observations. First, cross-lingual and low-resource settings continue to challenge even the most capable LLMs. Second, the HCE metric captures an essential dimension of linguistic modeling—calibration to human variability—that complements traditional accuracy measures. Models such as Compound-Beta and Gemma2-9B, which perform well on both axes, show promise for deployment in multilingual and semantically nuanced tasks.

\subsection{Reasoning Quality Assessment}
To evaluate whether large language models (LLMs) provide human-aligned justifications for their predictions, we conducted a reasoning quality assessment using semantic similarity metrics. Specifically, we compared the explanations generated by each LLM to ground truth rationales provided by expert annotators. We utilized the \texttt{MiniLM-L6-v2} model from the SentenceTransformers library to encode each explanation into high-dimensional embeddings, and computed the cosine similarity between the model-generated and ground-truth explanations. This similarity score served as a proxy for the semantic fidelity of the model's reasoning, where higher scores indicate stronger alignment with human judgment.

To further dissect the linguistic behavior of LLMs, we categorized prompts into several linguistically-motivated groups: \textbf{Same Tense Different Voice (STDV)}, \textbf{Different Tense Different Voice (DTDV)}, \textbf{Different Tense Same Voice (DTSV)}, \textbf{Negation- Affirmative (NA)}, \textbf{Ambiguous}, and \textbf{Bengali}. This grouping enables a fine-grained analysis of how well models generalize their reasoning strategies under syntactic or semantic variations. As shown in Table~\ref{tab:reasoning_quality}, we observe considerable variance in performance across both models and prompt types. Notably, \textit{LLaMA-4-Scout-17B} demonstrated the highest overall similarity score (0.2200), indicating robust and generally consistent reasoning across categories. \textit{Compound-Beta} also performed competitively, particularly in ambiguous contexts (0.3970) and with same tense pairs (0.2278). In contrast, \textit{Gemma2-9B} struggled significantly with temporally shifted prompts (0.0799 in \texttt{DTDV}), suggesting difficulty in modeling tense-dependent semantic equivalence.

Across all models, performance on negation remains notably poor (e.g., LLaMA-3.3-70B scores 0.0060), reinforcing prior observations that negation understanding remains a persistent challenge for LLMs. Surprisingly, reasoning on Bengali prompts was relatively strong (e.g., LLaMA-4-Scout scored 0.3317), indicating that multilingual pretraining can help mitigate reasoning loss in non-English contexts. These findings underscore the importance of evaluating explanation quality across nuanced linguistic axes, as high task accuracy may obscure underlying reasoning inconsistencies. For multimedia applications that depend on language as a grounding modality such as retrieval, summarization, or human-computer interaction, the ability of an LLM to provide faithful and consistent reasoning is not merely desirable but essential.
\begin{figure*}
    \centering
    \includegraphics[width=1\linewidth]{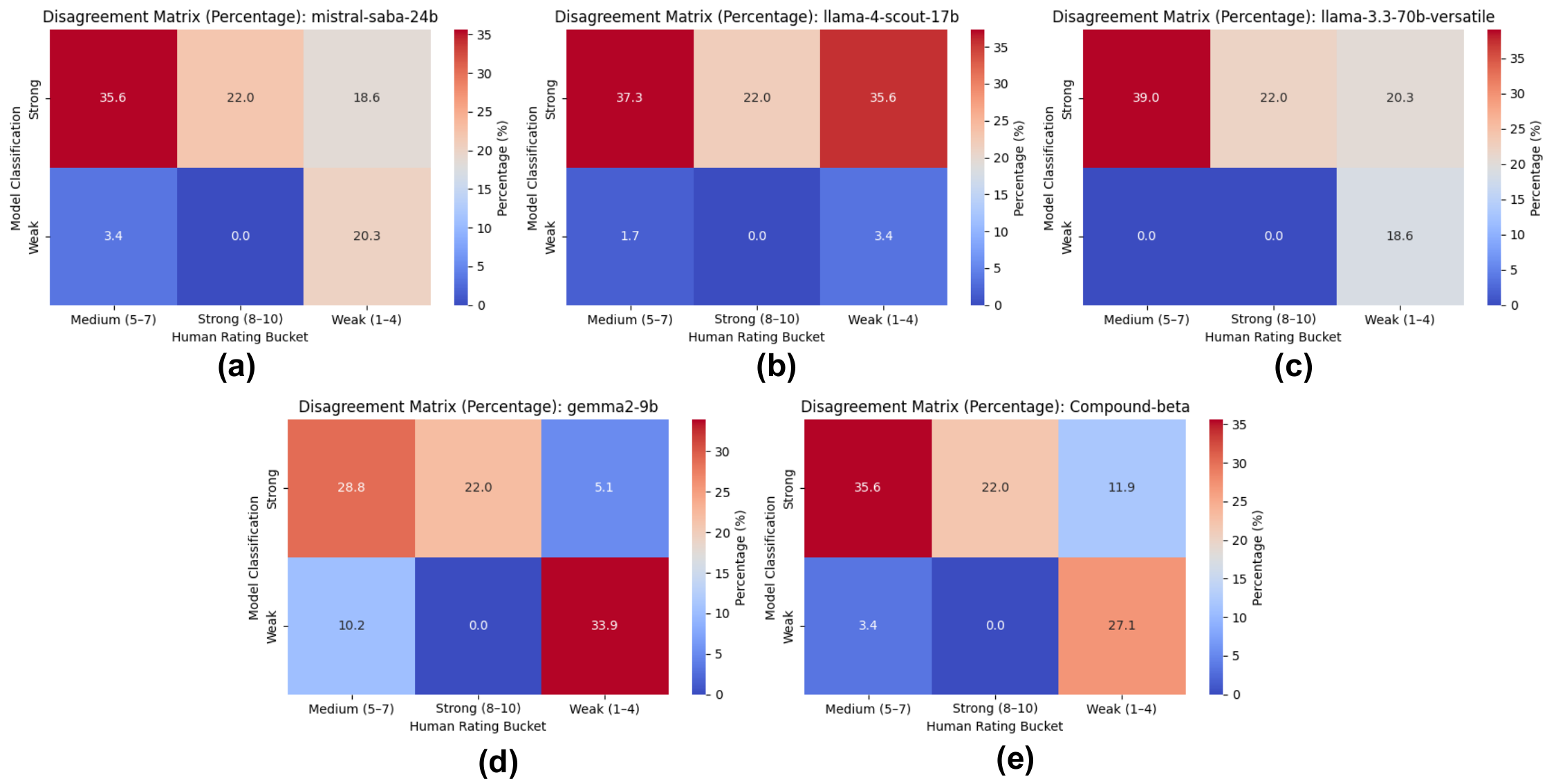}
    \caption{Disagreement matrices (in \%) between human rating buckets and model classifications for (a) Mistral-saba-24B (b) llama-4-scout-17B (c) llama-3.3-70B-Versatile (d) Gemma2-9B (e) Compound-Beta. Human ratings are grouped into three buckets—Strong (8–10), Medium (5–7), and Weak (1–4)—based on the median score. Model outputs are binarized into “Strong” ($\ge$8) and “Weak” ($\le$8). Each cell shows the percentage of total prompts falling into the corresponding model-human category. Higher percentages in off-diagonal cells indicate greater disagreement.}
    \label{disagree}
\end{figure*}
\subsection{Human Model Disagreement Analysis}

To systematically evaluate the alignment between model-generated assessments and human-annotated ratings, we construct \textit{Disagreement Matrices} for each language model. These matrices quantify the percentage distribution of prompts across human-defined quality buckets—\textbf{Strong (8--10)}, \textbf{Medium (5--7)}, and \textbf{Weak (1--4)}—in relation to a binary classification produced by the model (\textbf{Strong} if the model score $\geq$ 8; \textbf{Weak} otherwise). Each matrix cell reflects the proportion of prompts falling into the corresponding human-model label combination, providing a granular view of model agreement and deviation patterns.

Among the evaluated models, \textbf{LLaMA-3.3-70B-Versatile} demonstrates relatively strong alignment with human ratings in the \textit{Medium} category, covering 39.0\% of samples. However, it shows notable disagreement in the \textit{Weak--Strong} quadrant (20.3\%), where prompts rated poorly by humans are confidently scored as strong by the model, indicating an overestimation tendency. Interestingly, this model never classified a prompt as \textit{Weak} when humans rated it \textit{Medium} or \textit{Strong}, revealing a one-sided bias in its disagreement. \textbf{LLaMA-4-Scout-17B} follows a similar behavioral pattern but with an even broader disagreement footprint. A significant 35.6\% of prompts rated \textit{Weak} by human annotators were misclassified as \textit{Strong}, reflecting a consistent overconfidence in the model’s judgments.

In contrast, \textbf{Gemma2-9B} exhibits disagreement in the opposite direction. While it aligns well with humans in the \textit{Strong--Strong} quadrant (22.0\%), it underestimates quality in many cases: 10.2\% of \textit{Medium}-rated prompts were labeled \textit{Weak} by the model, and 33.9\% of prompts aligned in the \textit{Weak--Weak} cell, suggesting a conservative bias. \textbf{Compound-Beta} presents a more balanced disagreement profile. It maintains moderate agreement across all human rating buckets, with 27.1\% of prompts jointly labeled \textit{Weak} by both human raters and the model, indicating reliable performance in identifying low-quality content. However, minor discrepancies remain, particularly in the \textit{Medium} region.

\textbf{Mistral-Saba-24B} demonstrates a comparable profile to Compound-Beta but with slightly increased ambiguity. It records 20.3\% agreement in the \textit{Weak \& Weak} cell and 18.6\% disagreement in the \textit{Strong \& Weak} quadrant, pointing to occasional misclassification of high-quality prompts. Collectively, these disagreement matrices reveal not only the \textbf{degree of alignment} between model and human ratings but also the \textbf{qualitative nature of divergences}. Overconfidence in weak content (as seen in LLaMA models) and underestimation of mid-quality inputs (as in Gemma2-9B) highlight the importance of disagreement-aware evaluation frameworks. 
\section{Conclusion}

In this article, we have systematically evaluated the performance of state-of-the-art large language models (LLMs) in understanding subtle yet critical linguistic nuances, including negation, tense, voice, and modal variations, across both English and Bengali. Leveraging the PRECISE guidelines, we designed a rigorous framework for this evaluation, incorporating both traditional performance metrics and the novel Human Calibration Envelope (HCE) accuracy, which captures how closely model predictions align with human variability.

Our results reveal notable differences in how various models handle these linguistic complexities. \textbf{Compound-Beta} consistently demonstrated balanced and reliable performance across all tasks and languages, making it a strong candidate for applications requiring nuanced semantic understanding. \textbf{Gemma2-9B} showed exceptional alignment with human calibration in English, while \textbf{LLaMA-3.3-70B-Versatile} excelled in mixed and English scenarios but faltered in Bengali contexts. In contrast, \textbf{Mistral-SABA-24B} and \textbf{LLaMA-4-Scout-17B} offered more generalized performance without leading in any specific metric.

Our reasoning quality assessment further illuminated the models' capacity to generate human-aligned explanations. While certain models, such as \textbf{LLaMA-4-Scout-17B}, exhibited robust and consistent reasoning across prompt types, persistent challenges remain, particularly in understanding negation and temporally shifted language. Interestingly, some models performed better in Bengali reasoning tasks, suggesting that multilingual pretraining may offer partial mitigation of these challenges.

These findings have critical implications for multimedia applications that rely on language as a grounding modality—such as retrieval systems, video summarization, and interactive media generation. In these contexts, the ability of LLMs to model subtle linguistic cues and produce human-aligned reasoning is not merely beneficial, but essential for maintaining coherence, interpretability, and trust in multimedia interactions.

Taken together, these findings underscore that despite recent advancements, LLMs still face significant hurdles in capturing the fine-grained nuances of human language—especially in low-resource or morphologically rich languages. This work highlights the need for future research to focus not only on improving overall task performance but also on ensuring models are calibrated to human-like interpretive flexibility. Such efforts are critical for the deployment of LLMs in real-world scenarios that demand robust, multilingual, and semantically grounded interactions.

While our current evaluation emphasizes alignment with human similarity ratings and calibration to inter-annotator variance, future work will extend this analysis toward deeper interpretability and behavioral diagnostics of LLMs. Specifically, we plan to incorporate \textit{explainability methods} such as LIME, SHAP, and Integrated Gradients to probe whether the model’s attention to lexical tokens corresponds with human-perceived semantic pivots. Additionally, we aim to quantify \textit{inter-model agreement} using metrics like Krippendorff’s Alpha and Cohen’s Kappa to assess whether diverse LLMs interpret prompts consistently and whether such consensus aligns with human judgments. To further stress-test semantic sensitivity, we propose constructing \textit{contrastive minimal pairs} prompt pairs that differ only in subtle semantic or syntactic dimensions (e.g., tense shifts) and evaluating if models appropriately distinguish between them. For example, differentiating between "She ate the cake" and "She had eaten the cake" and their respective passive forms would reveal sensitivity to temporal semantics. Ideally, models exhibiting true linguistic competence should rank such pairs distinctly, in line with human ratings.

Lastly, we aim to explore \textit{uncertainty quantification} by leveraging the inter-quartile range (IQR) of human ratings as a proxy for interpretive subjectivity. We hypothesize that if models meaningfully understand language, their output variance across trials or model variants should be lower in cases of strong human consensus and higher where human disagreement is pronounced. These directions offer a principled path toward evaluating LLMs not just on accuracy, but on interpretability, calibration, and linguistic fidelity.

\begin{acks}
Part of this research was performed in the DeepDuo Foundation AI\&ML Research Lab. 
\end{acks}

\bibliographystyle{ACM-Reference-Format}
\bibliography{sample-base}

\end{document}